\documentclass[10pt,twocolumn,letterpaper]{article}

\usepackage{cvpr}
\usepackage{times}
\usepackage{epsfig}
\usepackage{graphicx}
\usepackage{amsmath}
\usepackage{amssymb}
\usepackage{array}
\usepackage{pifont}
\usepackage{capt-of,etoolbox}
\usepackage{multirow}
\usepackage{cite}
\usepackage{hhline}
\usepackage{authblk}

\usepackage[pagebackref=true,breaklinks=true,letterpaper=true,colorlinks,bookmarks=false]{hyperref}

\cvprfinalcopy 

\def\cvprPaperID{****} 

\ifcvprfinal\fi
\begin{document}

\title{Geometry-Aware Instance Segmentation with Disparity Maps}

\author{Cho-Ying Wu$^1$ \: Xiaoyan Hu$^2$ \: Michael Happold$^2$\: Qiangeng Xu$^1$\: Ulrich Neumann$^1$\\
$^1$University of Southern California \:\: 
$^2$Argo AI\\
{\tt\small \{choyingw, qiangenx, uneumann\}@usc.edu \:\: \{xiaoyan, mhappold\}@argo.ai }
\vspace{-5pt}
}

\makeatletter
\let\@oldmaketitle\@maketitle
\renewcommand{\@maketitle}{\@oldmaketitle
    \vspace{-14pt}
  \centering\includegraphics[width=0.86\linewidth]{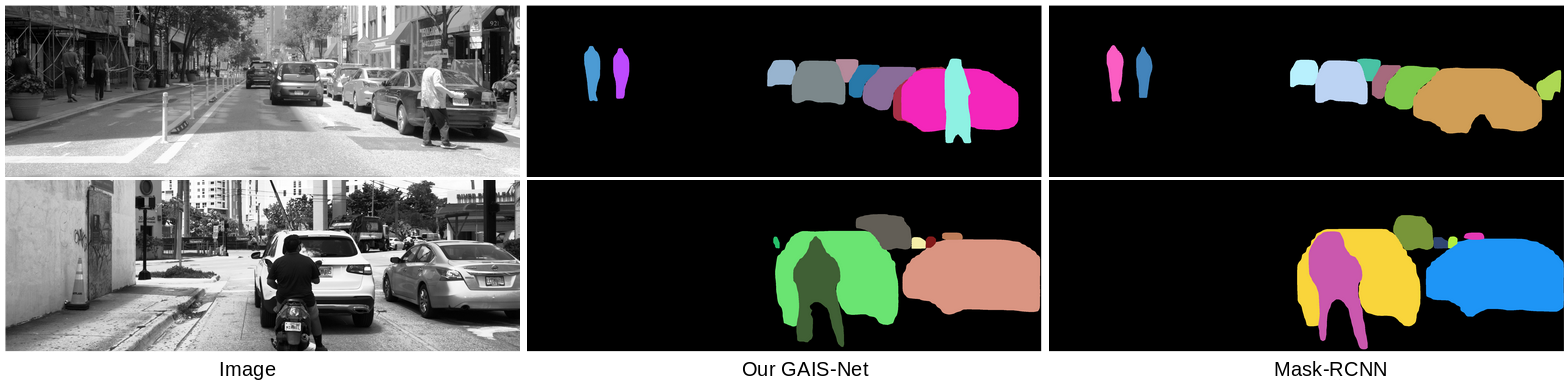}
  \captionof{figure}{\textbf{GAIS-Net results on HQDS dataset.} Left column shows stereo left images with histogram equalization to enhance contrast for better visualization. Middle and right column show Mask-RCNN and GAIS-Net results, respectively. Each instance has different colors. With the aid of geometric information, GAIS-Net can segment out the person from the overlapping area in the first row example. In the second row scenario, Mask-RCNN generates distorted mask for the smoking motorcyclist because of cigarette plume and in contrast GAIS-Net displays a more robust shape control capability.\newline}
  
  \label{pipeline}}
\makeatother

\maketitle

\begin{abstract}

\vspace{-10pt}
Most previous works of outdoor instance segmentation for images only use color information. We explore a novel direction of sensor fusion to exploit stereo cameras. Geometric information from disparities helps separate overlapping objects of the same or different classes. Moreover, geometric information penalizes region proposals with unlikely 3D shapes thus suppressing false positive detections. Mask regression is based on 2D, 2.5D, and 3D ROI using the pseudo-lidar and image-based representations. These mask predictions are fused by a mask scoring process. However, public datasets only adopt stereo systems with shorter baseline and focal legnth, which limit measuring ranges of stereo cameras. We collect and utilize High-Quality Driving Stereo (HQDS) dataset, using much longer baseline and focal length with higher resolution. Our performance attains state of the art. Please refer to our \href{http://www-scf.usc.edu/~choyingw/works/GAIS-Net/index.html}{project page} for codes and data. The full paper is available \href{https://drive.google.com/open?id=1ClWVotE94DfrjedJhZfDVHvp6WQ4aeMD}{here}.
\vspace{-12pt}
\end{abstract}

\section{Introduction}

Instance segmentation, which segments every object of interest, is an elemental task for computer vision. It is crucial for autonomous driving because it is vital to know positions of every object instance on roads. In the context of instance segmentation on images, previous approaches only operate on RGB imagery, such as Mask-RCNN \cite{he2017mask}. However, image data could be affected by illumination, color change, shadows, or optical defects. These factors can degrade the performance of image-based instance segmentation. By utilizing another modality that provides geometric cues of scenes~\cite{wu2023meta,wu2022toward,wu2023inspacetype,xu2020grid,wu2021scene}, and since object shapes are independent of object texture and color change, these strong priors add more robust information of the scenes. A prior work \cite{ye2017depth} that goes beyond the dominant paradigm to incorporate depth information only uses it for naive ordering rather than directly regressing masks or building an end-to-end trainable model to propagate depth information. Besides, their depth maps are predicted from monocular images, making the depth ordering unreliable.


In outdoor scenes, stereo cameras or lidar sensors are commonly used for depth acquisition. Stereo cameras are low-cost and their adjustable parameters, such as \textit{longer baselines} ($b$) and \textit{focal lengths} ($f$), favor stereo matching at far fields. Relation of depth and disparity is given by
\begin{equation}
\begin{aligned}
    depth = \frac{f\times b}{disparity}.
\end{aligned}
\label{dd_relationship}
\end{equation}
1-disparity (the minimal pixel difference showing the ideal longest range a stereo system could detect) represents farther distance if using longer \textit{f} and \textit{b}. Next, longer baselines and focal lengths favor more precise geometric estimations \cite{okutomi1993multiple}, since longer baselines produce smaller triangulation error, and longer focal lengths project objects on images with more pixels and thus enhance the robustness of stereo matching and show more complete shapes.

In this paper, we propose Geometry-Aware Instance Segmentation Network (GAIS-Net) that takes the advantages of both the semantic information from image domain and geometric information from disparity maps. Our contributions are summarized as follows:

1. To our knowledge, we are the first to perform instance segmentation on imagery by fusing images and disparity information to regress object masks.

2. We collect High-Quality Driving Stereo (HQDS) dataset, with a total of 8.8K stereo pairs and with $f \times b$ \textbf{4 times larger} than the current best dataset, Cityscapes.

3. We present GAIS-Net, an aggregation of representation design for instance segmentation using images, image-based, and point cloud-based networks. We train GAIS-Net with different losses, and fuse these predictions using the mask scoring. GAIS-Net achieves the state of the art.

\section{Method}

Our goal is to construct an end-to-end trainable network to perform instance segmentation for autonomous driving. Our system segments each instance and outputs confidence scores for bounding boxes and masks for each instance. To exploit geometric information, we adopt PSMNet \cite{chang2018pyramid}, the state-of-the-art stereo matching network, and introduce disparity information at ROI heads. The whole network design is in Fig. \ref{pipeline}.

\begin{figure*}[bt!]
\begin{center}
\includegraphics[width=1.0\linewidth]{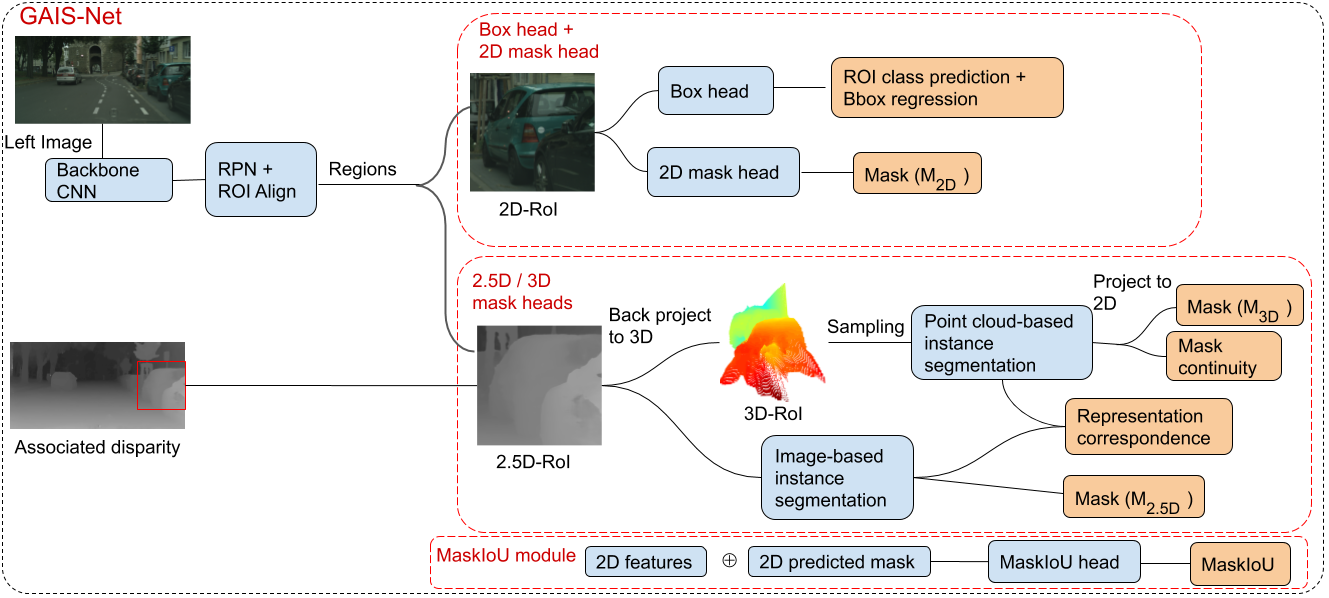}
\end{center}
\vspace{-15pt}
   \caption{\textbf{Network design of our GAIS-Net.} Bbox is for bounding box. We color modules in blue and outputs or loss parts in orange. In the MaskIoU module, the 2D features and 2D predicted mask are from the 2D mask head. They are fed into MaskIoU head to regress MaskIoU scores. We draw the MaskIoU head separately for clear visualization. $\oplus$ stands for concatenation.}
\label{pipeline}
\end{figure*}

We build a two-stage detector with a backbone network, such as ResNet50-FPN, and a region proposal network (RPN) with non-maximum suppression. Object proposals are collected by feeding a stereo left image into the backbone network and RPN. The same as Mask-RCNN, we perform bounding box regression, class prediction for proposals, and mask prediction based on image domain features. Corresponding losses are denoted as $\mathcal{L}_{box}$, $\mathcal{L}_{cls}$, and $\mathcal{L}_{2Dmask}$, and are identified in \cite{he2017mask}.

\subsection{Geometry-Aware Mask Prediction}
\label{3_1}
\textbf{2.5D ROI and 3D ROI.}\hspace{0.2cm} We use PSMNet \cite{chang2018pyramid} and stereo pairs to predict dense disparity maps, projected onto the left stereo frame. Next, RPN outputs region proposals. We collect proposals and crop out these areas from the disparity map. We call these cropped out disparity areas as \textit{2.5D ROI}.

Based on the observations from pseudo-lidar work \cite{wang2019pseudo}, which describes the advantage of back-projecting 2D grid structured data into 3D point cloud and processing with point cloud networks, we back-project the disparity map into $\mathbb{R}^3$ space, where for each point, the first and second components describe its 2D grid coordinates, and the third component stores its disparity value. We name this representation as \textit{3D ROI}.


\textbf{Instance Segmentation Networks.} \hspace{0.2cm}Each 3D ROI contains different number of points. To facilitate training, we uniformly sample the 3D ROI to 1024 points, and collect all the 3D ROI into a tensor. We develop a PointNet structured instance segmentation network to extract point features and perform per-point mask probability prediction. We re-project the 3D feature onto the 2D grid to calculate the mask prediction and its loss $\mathcal{L}_{3Dmask}$. The re-projection is efficient because we do not break the point order in the point cloud-based instance segmentation. $\mathcal{L}_{3Dmask}$, same as $\mathcal{L}_{2Dmask}$, is a cross-entropy loss between a predicted probability mask and its matched groundtruth.

To fully utilize advantages of different representations, we further do 2.5D ROI instance segmentation with an image-based CNN. Similar to instance segmentation on 2D ROI, this network extracts local features of 2.5D ROI, and later performs per-pixel mask probability prediction. The mask prediction loss is denoted as $\mathcal{L}_{2.5Dmask}$. 


\subsection{Mask Continuity}

\begin{figure}[bt!]
    \centering
    \includegraphics[scale=0.20]{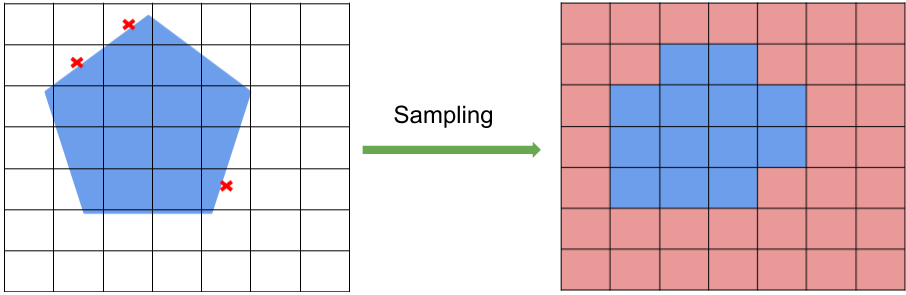}
    \vspace{-2pt}
    \caption{\textbf{Undesirable sampling example.} The blue areas represent foreground. Suppose we uniformly sample every grid center point in the left figure, resulting in the point cloud showing in the occupancy grid on the right. Red crosses are undesirable sampling points, which just lie outside the foreground object, making the shape after sampling different from the original one.}
    \vspace{-10pt}
    \label{sample}
\end{figure}

We sample 3D ROI to 1024 points uniformly. However, predicted masks, denoted as $M_{3D}$, and their outlines are sensitive to pseudo-lidar sampling strategies. An undesirable sampling is illustrated in Fig. \ref{sample}. To compensate for the undesirable effect, we introduce a mask continuity loss. Since objects are structured and continuous, we calculate a \textit{mask Laplacian} as $\nabla^2 M = \frac{\partial^2 M}{\partial x^2}+\frac{\partial^2 M}{\partial y^2}$, where $x$ and $y$ denote the dimensions of $M$. Mask Laplacian computes continuity of $M$. Further, the mask continuity loss is calculated as $\mathcal{L}_{cont}=\|\nabla^2 M\|^2$ for penalizing discontinuities of $M$.

\subsection{Representation Correspondence}

We use the point cloud-based network and the image-based network to extract features and regress $M_{3D}$ and $M_{2.5D}$. These two masks should be similar because they are from the same disparity map. To evaluate the similarity, cross-entropy is calculated between $M_{3D}$ and $M_{2.5D}$, and serves as a \textit{self-supervised} correspondence loss $\mathcal{L}_{corr}$.  Minimizing this term lets the networks of different representations supervise each other to extract more descriptive features for mask regressing, resulting in similar probability distribution between $M_{2.5D}$ and $M_{3D}$. Mask-RCNN uses a $14\times 14$ feature grid after ROI pooling to regress masks. We also use this size at the mask heads.




\subsection{Mask Scores and Mask Fusion}

MS-RCNN \cite{huang2019mask} introduces mask scoring to directly regress MaskIoU score based on a predicted mask and its associated matched groundtruth, showing quality of the mask prediction. However, their scores are not adopted at inference time to help manipulate mask shapes.

\begin{figure*}[bt!]
\begin{center}
\includegraphics[width=0.85\linewidth]{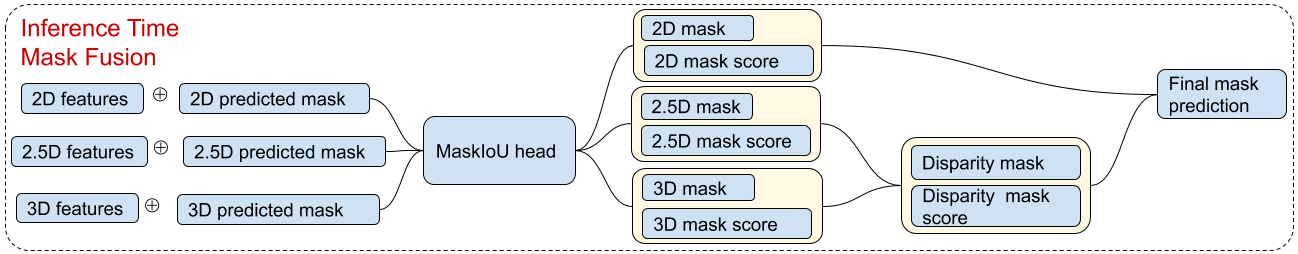}
\end{center}
\vspace{-14pt}
   \caption{\textbf{Inference time mask fusion of predictions from different representations.} We fuse the 2.5D mask and 3D mask first because they are from the same source. We then fuse the mask predictions from the image domain and disparity. $\oplus$ represents concatenation.}
\label{inference}
\vspace{-8pt}
\end{figure*}

We adopt mask scoring and further exploit MaskIoU scores to fuse mask predictions from different representations at the inference time. The mask fusion process is illustrated in Fig. \ref{inference}. During the inference time, we concatenate features and predicted masks of different representations respectively as inputs to the MaskIoU head. Scores of $s_{2D}, s_{2.5D},$ and $s_{3D}$ are outputs from the MaskIoU head. We fuse mask predictions using their corresponding mask scores. We first linearly combine ($M_{2.5D}$, $s_{2.5D}$) and ($M_{3D}$, $s_{3D}$) to obtain ($M_D$, $s_D$) for the disparity. The formulation is as follows.
\begin{equation}
     M_D=M_{2.5D}\times\frac{s_{2.5D}}{s_{2.5D}+s_{3D}}+M_{3D}\times\frac{s_{3D}}{s_{2.5D}+s_{3D}},
\end{equation}
\begin{equation}
     s_D= s_{2.5D}\times\frac{s_{2.5D}}{s_{2.5D}+s_{3D}} + s_{3D}\times\frac{s_{3D}}{s_{2.5D}+s_{3D}}.
\end{equation}
Later, we linearly fuse ($M_{2D}, s_{2D}$) and ($M_D, s_D$) likewise to obtain the final probability mask $M_f$ and its corresponding final mask score. The inferred mask is created by binarizing $M_f$.

The mask scoring process should not be different for each representation. We only use 2D image features and $M_{2D}$ to train a single MaskIoU head instead of constructing 3 MaskIoU heads for each representation. In this way, the MaskIoU module would not add much more memory use and the training is also effective. The MaskIoU loss is denoted as $\mathcal{L}_{miou}$.


\section{Experiments}
\subsection{HQDS Dataset}
\label{hqds}
Outdoor RGBD scene understanding is still less explored since much longer range sensing is required to align information from images and depth. Such as vehicles at distances showing in images but undetected by a depth sensor would bring ambiguity into RGBD methods. To conduct exploration of outdoor RGBD methods, and provide high quality data to reveal advantages of sensor fusion, we collect High-Quality Driving Stereo (HQDS) dataset in urban environments. Table \ref{Dataset} shows a comparison with other public datasets for instance segmentation. Image resolution of HQDS is $1024 \times 3072$. From the table and Eq. \ref{dd_relationship}. HQDS has the largest $f$ $\times$ $b$. Measuring range by the configuration is up to \textbf{1650 meters} with 1-pixel disparity, which is only \textit{440} and \textit{350} for Cityscapes and KITTI. Note that produced disparity maps are computed by stereo matching methods, so actual working distances are associated with methods' robustness and image noise. However, \textit{longer baselines and focal lengths} still favor \textit{far-field stereo matching} since the former could show \textit{better geometry and more complete shapes} for objects at distances \cite{okutomi1993multiple}.

\begin{table}[bt!]
\begin{center}
\caption{\textbf{Comparison between collected HQDS and other public datasets for instance segmentation with stereo data.} Stereo pairs \# means number of training stereo pairs. Stereo camera baseline ($b$) is in meters. $f_x$ is for horizontal focal length.}
\begin{tabular}{|p{1.3cm}<{\centering}||p{1.9cm}<{\centering}|p{1.2cm}<{\centering}|p{0.5cm}<{\centering}|p{1.1cm}<{\centering}|}
\hline
Dataset &  Resolution (megapixels) & Stereo Pairs \#  & $b$ (m) & $f_x$ (pixels) \\
\hline
Cityscapes &  2.09  & 2.7K & 0.2 & 2.2K \\
KITTI  &  0.71 & 0.2K & 0.5 & 0.7K \\
\textbf{HQDS}  &  \textbf{3.15} & \textbf{6K} & \textbf{0.5} & \textbf{3.3K}\\
\hline
\end{tabular}
\end{center}
\vspace{-20pt}
\label{Dataset}
\end{table}

HQDS contains 6K/1.2K stereo pairs for training/testing. We follow a half-automation process to annotate data with a group of supervised annotators. Our internal large-scale labeling system produces preliminaries, and the annotators adjust yielded bounding boxes and mask shapes or filter out false predictions to produce HQDS groundtruth.

There are 60K instances in the training set and 11K in the testing set. We adopt 3 instance classes: human, bicycle/motorcycle, and vehicle. Although other datasets on driving adopt more, such as Cityscapes use 8 classes, from MaskRCNN's study \cite{he2017mask} they suffer from much inter-class ambiguity which leads to biased results.

Associated number of instances in the training and testing sets are (5.5K, 1.5K, 52.8K) and (2.4K, 1K, 8.4K) respectively. Most of non-synthetic datasets encounter class-imbalanced issue. To remedy the imbalance, we adopt COCO dataset (instance segmentation for common objects) pretrained weights with class pruning in our implementations and comparison methods.



\textbf{Evaluation and Metrics.} We fairly compare with recent state-of-the-art methods validated on large-scale COCO dataset, including Mask-RCNN \cite{he2017mask}, MS-RCNN \cite{huang2019mask}, Cascade Mask RCNN \cite{chen2019hybrid}, and HTC \cite{chen2019hybrid} (w/o semantics), by using their publicly released codes and their COCO pretrained weights. We follow their training procedures to conduct comparison experiments.

We report numerical results in the \textbf{standard COCO-style}. Average precision (AP) averages across \textit{different IoU levels}, from 0.5 to 0.95 with 0.05 as an interval. AP$_{50}$ and AP$_{75}$ are 2 typical IoU levels. The units are \%. Table \ref{test1} shows the comparison with others. The proposed GAIS-Net attains the state of the art. We exceed Mask-RCNN using the same backbone by 9.7\% and 6.8\% for bounding box and mask AP, respectively.


\begin{table}[tb!]
\begin{center}
  \caption{\textbf{Quantitative comparison on HQDS testing set.} The first table is for bounding box evaluation. The second table is for mask evaluation.}
  \label{test1}
  \begin{tabular}{|c||
  p{0.6cm}<{\centering}|
  p{0.6cm}<{\centering}|
  p{0.6cm}<{\centering}|
  p{0.6cm}<{\centering}|
  p{0.6cm}<{\centering}|}
  \hline
    Bbox Evaluation  &  AP & AP$_{50}$ & AP$_{75}$ & AP$_{S}$ & AP$_{L}$ \\
    \hline
    Mask-RCNN &  36.3 & 57.4 & 38.8 & 19.1 & 51.9 \\ 
    MS-RCNN &  42.2 & 65.1 & 46.6 & 20.8 & 59.6 \\
    Cas. Mask-RCNN &  37.4 & 55.8 & 38.9 & 18.0 & 54.7 \\
    HTC &  39.4 & 58.3 & 43.1 & 18.5 & 57.9 \\
    GAIS-Net & \textbf{46.0} & \textbf{67.7} & \textbf{53.3} & \textbf{23.6} & \textbf{66.2} \\
    \hline
    \hline
    Mask Evaluation  &  AP & AP$_{50}$ & AP$_{75}$ & AP$_{S}$ & AP$_{L}$ \\
    \hline
    Mask-RCNN &  33.9 & 53.2 & 35.5 & 14.4 & 49.7 \\ 
    MS-RCNN &  39.2 & 61.3 & 40.4 & \textbf{18.8} & 56.4 \\
    Cas. Mask-RCNN &  33.4 & 54.4 & 34.8 & 11.7 & 49.5 \\
    HTC &  34.5 & 56.9 & 36.7 & 11.6 & 52.0 \\
    GAIS-Net &  \textbf{40.7} & \textbf{65.9} & \textbf{43.5} & 18.3 & \textbf{59.2} \\
    \hline
  \end{tabular}
  \vspace{-18pt}
\end{center}
\end{table}

\subsection{Cityscapes Dataset}
We also conduct experiments on Cityscapes dataset. However, its baseline and focal length are shorter than HQDS, and the maximal measuring distance is only \textbf{1/4} of HQDS. Much shorter focal length and baseline \textbf{limit the working distance} of stereo matching and produce disparity maps only focusing at near fields with \textit{poor shapes and geometry} \cite{okutomi1993multiple}. From Table \ref{cityscapes}, performance of GAIS-Net is still better than Mask-RCNN. The improvement gap between HQDS and Cityscapes is mainly caused by the latter's shorter baseline and focal length. 


\begin{table}[tb!]
\begin{center}
  \caption{\textbf{Instance segmentation results on Cityscapes datset.}}
  \label{cityscapes}
  \small
  \begin{tabular}{|p{2.8cm}<{\centering}|
  p{1.8cm}|
  p{1.3cm}<{\centering}|
  }
  \hline
    Evaluation & Training data &  Mask AP 
     \\
    \hline
    Mask-RCNN \cite{he2017mask} & fine only  & 31.5\\
    Our GAIS-Net & fine only  & \textbf{32.5}\\
    Mask-RCNN \cite{he2017mask} & fine + COCO  & 36.4\\
    Our GAIS-Net & fine + COCO & \textbf{37.1}\\
    \hline
  \end{tabular}
  \vspace{-18pt}
\end{center}
\end{table}

{\small
\bibliographystyle{ieee_fullname}
\bibliography{egbib}
}

\end{document}